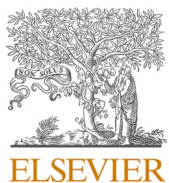
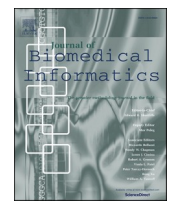

# Advancing Chinese biomedical text mining with community challenges

Hui Zong [a,1], Rongrong Wu [a,1], Jiaxue Cha [b], Weizhe Feng [a], Erman Wu [a], Jiakun Li [a,c], Aibin Shao [a], Liang Tao [d], Zuofeng Li [e], Buzhou Tang [f], Bairong Shen [a,*]

[a] *Joint Laboratory of Artificial Intelligence for Critical Care Medicine, Department of Critical Care Medicine and Institutes for Systems Genetics, Frontiers Science Center for Disease-related Molecular Network, West China Hospital, Sichuan University, Chengdu 610041, China*
[b] *Shanghai Key Laboratory of Signaling and Disease Research, Laboratory of Receptor-Based Bio-Medicine, Collaborative Innovation Center for Brain Science, School of Life Sciences and Technology, Tongji University, Shanghai 200092, China*
[c] *Department of Urology, West China Hospital, Sichuan University, Chengdu 610041, China*
[d] *Faculty of Business Information, Shanghai Business School, Shanghai 201400, China*
[e] *Takeda Co. Ltd., Shanghai 200040, China*
[f] *Department of Computer Science, Harbin Institute of Technology, Shenzhen 518055, China*

ABSTRACT

*Objective:* This study aims to review the recent advances in community challenges for biomedical text mining in China.
*Methods:* We collected information of evaluation tasks released in community challenges of biomedical text mining, including task description, dataset description, data source, task type and related links. A systematic summary and comparative analysis were conducted on various biomedical natural language processing tasks, such as named entity recognition, entity normalization, attribute extraction, relation extraction, event extraction, text classification, text similarity, knowledge graph construction, question answering, text generation, and large language model evaluation.
*Results:* We identified 39 evaluation tasks from 6 community challenges that spanned from 2017 to 2023. Our analysis revealed the diverse range of evaluation task types and data sources in biomedical text mining. We explored the potential clinical applications of these community challenge tasks from a translational biomedical informatics perspective. We compared with their English counterparts, and discussed the contributions, limitations, lessons and guidelines of these community challenges, while highlighting future directions in the era of large language models.
*Conclusion:* Community challenge evaluation competitions have played a crucial role in promoting technology innovation and fostering interdisciplinary collaboration in the field of biomedical text mining. These challenges provide valuable platforms for researchers to develop state-of-the-art solutions.

## 1. Introduction

Over the past few decades, the field of biomedical research has witnessed a remarkable growth in the accumulation of extensive amounts of textual data[1–4]. These data come from various sources with a substantial volume, including scientific literatures, electronic health records, clinical trial reports, social media platforms, books, patents, and more. These data contain rich information that can be leveraged for knowledge discovery [5,6], hypothesis generation[7,8] and clinical practice[9,10]. However, due to the vast amount and complexity of textual resources, manual reading and processing of data are time-consuming, labor-intensive and inefficient. Researchers and clinicians are facing the challenges of information explosion and knowledge emergence. As a result, there is a need in the development of efficient computational techniques for health information processing.

Biomedical text mining, also known as biomedical natural language processing (BioNLP) has gained significant attention [11–15]. BioNLP can extract key biological entities (such as variant, gene, protein, and disease) [16–19] and medical entities (such as treatment, surgery, and drug) [20–23], identify entity relationships [24–27], perform document classification [28,29], information retrieval [30], and knowledge question answering [31], among other tasks. BioNLP techniques have found extensive applications in scientific research and clinical practice. For instance, the identification of new drug targets [32], the discovery of novel therapeutic interventions [33–35], the exploration of adverse drug reactions [36], and the cohort building of clinical trials [37–39]. BioNLP techniques can also facilitate the construction of knowledge bases and ontologies [40–43], enabling efficient data integration and interoperability across different sources [44]. Furthermore, cutting-edge large language models have demonstrated remarkable applications in the fields of biomedical research and healthcare [45–47].

Community challenges have emerged as crucial catalysts for promoting technological innovation and interdisciplinary collaboration in the field of BioNLP. These challenges provide platforms for researchers, data scientists, and domain experts to exhibit their skills, exchange innovative ideas, and develop cutting-edge solutions for data mining and information processing in biomedical research [48]. By providing standard datasets manually annotated by domain experts and specifically designed evaluation tasks, BioNLP community challenge evaluations foster the development of robust algorithms, novel methodologies, and benchmark frameworks. Over the past few decades, multiple renowned community challenges, including BioCreative [25,29], TREC [49], and i2b2 [38,50], have made significant contributions to advancing biomedical text mining technology.

In recent years, China has organized numerous biomedical text mining challenges aimed at solving problems specific to Chinese biomedical and health information processing [51–56]. These challenges have yielded valuable insight and advancements in the understanding of Chinese text and the processing of health information in the Chinese context. However, there remains a noticeable gap in systematic summaries and comparative analyses of these community challenges. In addition, with the rapid development of large language models such as ChatGPT, the future organization and attention of Chinese BioNLP community challenges face new opportunities and challenges.

This review aims to provide a comprehensive landscape for the recent advances in community challenges of Chinese biomedical text mining. We first collect evaluation shared tasks organized by academic conference, and conduct systematic summary and comparative analysis for specific tasks, including data sources and task types. Then, we summarize the potential clinical applications of these community challenge tasks from translational informatics perspective. Finally, we compare with their English counterparts, and discuss the contributions, limitations, lessons and guidelines of these community challenges tasks, while highlighting future directions in the era of large language models.

| | |
|---|---|
| Problem or Issue | The growth of biomedical NLP in China requires a comprehensive understanding of recent community challenges to guide future developments, especially in the era of large language models. |
| What is Already Known | China has organized numerous biomedical NLP challenges, providing valuable insights into Chinese biomedical and health information processing. However, systematic summaries and comparisons of these challenges are lacking. |
| What this Paper Adds | This review provides a comprehensive overview of recent advances in Chinese biomedical NLP challenges, including systematical summaries, clinical applications, comparisons with English counterparts, contributions, limitations, lessons learned, proposed guidelines, as well as future directions in the context of emerging large language models. |

## 2. Community challenges overview

Fig. 1 presents the timeline overview of these challenges spanning the years 2017 to 2023. Each community challenge is represented by a different background color. The challenge names are presented in white color, while the specific shared tasks within each challenge are shown in black. The challenge tasks were initially introduced to the Chinese biomedical text mining community by China Conference on Knowledge Graph and Semantic Computing (CCKS) in 2017, and was subsequently gained prominence through the China Health Information Processing Conference (CHIP). Others such as Chinese Conference on Information Retrieval (CCIR), Chinese Society of Medical Information (CSMI), China National Conference on Computational Linguistics (CCL), and Digital China Innovation Contest (DCIC) also contributed in recent years.

### 2.1. CCKS

As shown in Table 1, in 2017, CCKS held first Chinese Biomedical Text Mining Evaluation, focusing on named entity recognition (NER) in electronic medical records [57]. As a basic natural language processing (NLP) task, CCKS has held clinical named entity recognition (CNER) task for five consecutive years from 2017 to 2021 [55–59]. In 2019, entity attribute recognition was additionally introduced [55], and in 2020, event extraction was introduced [56]. Furthermore, in 2020, knowledge graph construction and question-answering tasks specifically related to COVID-19 were organized [60]. In 2021, CCKS expanded its scope to include various task types such as medical entity recognition and event extraction, link prediction in a multi-level knowledge graph involving phenotypes, drugs, and molecules, generation of medical dialogues containing implicit entities, and reading comprehension of medical popular science knowledge [59,61,62]. In 2023, based on the Chinese Biomedical Language Understanding Evaluation (CBLUE) dataset, CCKS transformed 16 different NLP tasks in medical scenarios into prompt-based language generation tasks, establishing the first Chinese benchmark for evaluating language models in the medical domain [63,64]. It is worth noting that the evaluation tasks organized in the CCKS community challenge encompass a broad range of domains, with biomedical and healthcare being just one of them.

### 2.2. CHIP

CHIP, as an organization specialized in the processing of health information in Chinese, released multiple community challenge evaluation tasks focused on biomedical text mining each year (Table 2). In 2018, two tasks were released, namely medical entity recognition and attribute extraction, as well as health consultation question pairs matching. In 2019, three tasks were released, including clinical terminology standardization, disease question pairs similarity calculation task and clinical trial eligibility criteria text classification task [28]. In 2020, six tasks were released [65,66]. For task types, literature-based question generation and COVID-19 trends prediction are new added. For data types, in addition to electronic medical records, medication instructions and traditional Chinese medicine (TCM) literature were also involved. In 2021, three tasks were released. It is worth noting that in 2021, CHIP introduced the CBLUE [67], which encompassed the 8 biomedical NLP tasks as benchmark subtasks. In 2022, five tasks were released [51–54,68]. For the first time, CHIP included tasks related to optical character recognition and information extraction from medical paper documents [68]. Additionally, CBLUE released its second version, which comprised 18 subtasks. In 2023, six tasks were released, two of which were related to the evaluation of medical large language models [69–74]. One dataset originated from CBLUE [67], while the other dataset was created manually.

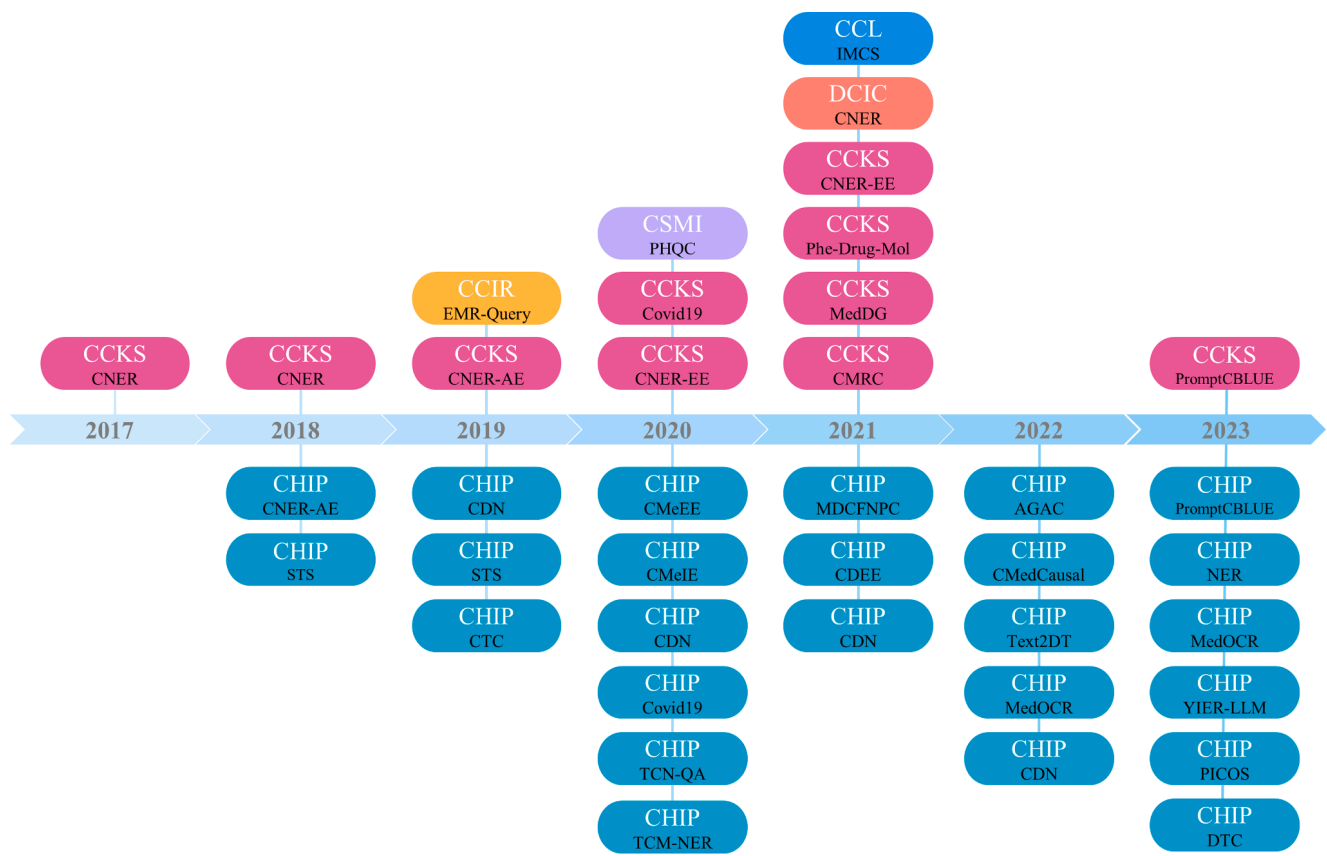

**Fig. 1.** Timeline overview of community challenges in Chinese biomedical text mining. Different challenges are shown by different background colors. The challenge names are shown in white, while the specific tasks within each challenge are shown in black. CCKS: China Conference on Knowledge Graph and Semantic Computing, CHIP: China Health Information Processing Conference, CCIR: Chinese Conference on Information Retrieval, CSMI: Chinese Society of Medical Information, CCL: China National Conference on Computational Linguistics, DCIC: Digital China Innovation Contest.

### 2.3. CCIR, CSMI, CCL and DCIC

There are some other community challenges that occasionally released biomedical text mining evaluation tasks (Table 3). For example, in 2019, CCIR introduced a task, which required system to return results given a medical event knowledge graph and a series of natural language questions. In 2020, CSMI launched a task for classifying public health questions into 6 categories. In 2021, CCL released an intelligent medical dialogue diagnosis and treatment task, which contained 3 subtasks, including the extraction of medical entities and symptom information from medical dialogue texts, the generation of structured medical reports, and the simulation of dialogues process to determine specific diseases.

### 3. Evaluation tasks overview

Fig. 2 presents the distribution of data sources, organizations, and artificial intelligence tasks in the Chinese biomedical text mining community challenge. For data sources, we categorize into electronic health records (EHRs) and non-electronic health records (non-EHRs). EHR comprises various types such as radiology reports, pathology reports, and scanned documents of paper-based medical records. Non-EHR consists of published literatures (e.g., title and abstract text), internet medical data (e.g., health consultation questions, medical popularization content, online consultation data, and doctor-patient question-answering dialogues), clinical experiment registration documents (e.g., eligibility criteria text), clinical practice guidelines, medical textbooks, and medication instructions. It is important to note that most tasks used textual data from a single source, with only a few tasks incorporating both EHR and non-EHR data. For task organization, the results indicate that the majority of tasks are jointly organized by researchers from both academia and industry, highlighting the intersection and collaboration between the fields of medicine and engineering. In addition, the participation of medical artificial intelligence related companies can promote the integration of developed models into clinical systems, thus contributing to clinical informatics and healthcare delivery. The number of tasks organized by academic researchers is higher than industrial researchers, suggesting that academia plays a pivotal role in driving advancements within the field. For artificial intelligence tasks, NER is the most focused task, followed by other tasks including question answering, text classification, relationship extraction, entity normalization, knowledge graph, text similarity, event extraction, large language model evaluation, attribute extraction, optical character recognition, and text generation. In the following sections, we will provide detailed descriptions of these tasks.

**Table 1**
Summary of evaluation tasks in China Conference on Knowledge Graph and Semantic Computing (CCKS).

| Year | Evaluation task abbreviation | Evaluation task title | Dataset description | Dataset size | Link |
|---|---|---|---|---|---|
| 2023 | PromptCBLUE | CCKS-PromptCBLUE medical large model evaluation | The dataset is sourced from the CBLUE benchmark, encompassing 16 scenarios of medical natural language processing tasks, and includes 94 instruction fine-tuning templates. | Training set: 68500, validation set: 10270, test set: 20,540 | https://sigkg.cn/ccks2023/evaluation |
| 2021 | CNER-EE | Entity and event extraction of Chinese electronic medical records | The medical named entity recognition dataset consists of manually annotated plain text documents from EHRs, identifying medically relevant entities. It includes 6 predefined categories: diseases and diagnoses, examinations, tests, surgeries, medications, and anatomical locations. The medical event extraction dataset includes manually annotated plain text documents from EHRs, focusing on attribute entities related to primary entities of tumor events. It encompasses 3 categories: primary site, lesion size, and metastatic site. | 2800 and 3000 | https://sigkg.cn/ccks2021/?page_id=27 |
| 2021 | Phe-Drug-Mol | Link Prediction of Phenotype-Drug-Molecular Multilevel Knowledge Graph | A knowledge graph constructed from structured data sourced from reputable websites encompasses 7 types of relationships: associated_with, disease_mapped_to_gene, treats, targets, interacts_with, annotates, and pathway_has_gene_element. | Dataset size: 80,000 entities, 1,200,000 triples. | https://sigkg.cn/ccks2021/?page_id = 27 |
| 2021 | MedDG | Chinese Medical Dialogue Generation Incorporating Implicit Entities | The MedDG dataset, annotated with entities, encompasses 12 types of gastroenterology-related diseases. Each dialogue is annotated with 160 relevant entities across 5 categories: diseases, symptoms, attributes, examinations, and medications. Dataset size: 20,611. | Training set: 17864, test set: 4347 | https://sigkg.cn/ccks2021/?page_id=27 |
| 2021 | CMRC | Reading Comprehension for Chinese Medical Popular Science Knowledge | The dataset for reading comprehension of medical popular science knowledge includes the main content and a list of question–answer pairs, including question description, question ID, answer list. The dataset of recognizing irrelevant answers in medical popular science knowledge, the format of this dataset is one entry per line, with five columns, including Label, Docid, Question, Description, and Answer. | 36,000 and 55,000 | https://sigkg.cn/ccks2021/?page_id=27 |
| 2020 | Covid19 | Construction and Question-Answering of COVID-19 Knowledge Graph | The COVID-19 knowledge graph encompasses seven entity types: virus, bacteria, disease, drug, medical specialty, examination subject, and symptom. The COVID-19 concept graph additionally includes type relationships between entities and concepts, as well as hierarchical relationships among concepts. The antiviral drug graph includes entities, entity attributes, and relationships between entities. The integrated dataset from the open-domain knowledge base PKUBASE and the OpenKG COVID-19 special topic includes information on entity category triples, hierarchical relationships between types, and predicates. | This knowledge graph contains 66,499,920 triples, 25,574,536 entities, and 408,690 relations. A training set of 4,000 items, a validation set of 1,529 items, and a test set of 1,599 items | https://sigkg.cn/ccks2020/?page_id = 516 |
| 2020 | CNER-EE | Entity and event extraction of Chinese electronic medical records | For medical named entity recognition dataset, it consists of manually annotated entities in EHRs, including diseases and diagnoses, examinations, tests, surgeries, medications, and anatomical locations. For medical event extraction dataset, it comprises manually annotated attribute entities associated with primary entities of oncology events in EHRs. It includes three categories: primary site, lesion size, and metastatic site. | 1500 and 1400 | https://sigkg.cn/ccks2020/?page_id=516 |
| 2019 | CNER-AE | Named entity recognition and attribute extraction for electronic health records | Medical Named Entity Recognition Dataset: The dataset comprises manually annotated documents from EHRs, capturing clinically relevant entities. These entities are categorized into 5 pre-defined categories: Symptoms and Signs, Examinations and Tests, Diseases and Diagnoses, Treatments, and Body Parts. Dataset size: 12,020. Attribute Extraction Dataset: This dataset consists of manually annotated documents from EHRs, focusing on attribute entities related to tumor events. These entities are categorized into 3 types: Lesion Size, Primary Site, and Metastatic Site. Dataset size: 2,000. | 1379 | https://www.sigkg.cn/ccks2019/?page_id=62 |
| 2018 | CNER | Named entity recognition for electronic health records | The dataset consists of manually annotated entities from EHRs, including anatomical sites, symptom descriptions, independent symptoms, medications, and surgeries. | 800 | https://www.sigkg.cn/ccks2018/?page_id=16 |

Table 1 (*continued*)

| Year | Evaluation task abbreviation | Evaluation task title | Dataset description | Dataset size | Link |
| --- | --- | --- | --- | --- | --- |
| 2017 | CNER | Named entity recognition for electronic health records | The dataset consists of manually annotated entities from EHRs, including anatomical locations, symptom descriptions, independent symptoms, medications, and surgeries. | 400 | https://www.sigkg.cn/ccks2017/?page_id=51 |

### 3.1. Information extraction

Information extraction is a fundamental task in the field of biomedical text mining, encompassing the extraction of domain-specific entities, attributes, relationships, and events from both structured and unstructured biomedical texts.

NER is the most common task in Chinese biomedical text mining research. From 2017 to 2021, CCKS continuously organized CNER tasks. These task data were all derived from electronic medical records from real hospitals, with variations in entity types and dataset sizes each year. In 2017, the task defined five entity types, including symptom and sign, examination and test, disease and diagnosis, treatment, and body part [57]. In 2018, the task defined five entity types, including anatomical site, symptom description, symptoms item, medication, and surgery [58]. In 2019, the task defined the same five entity types as 2017 [55]. In 2020, six entity types were defined, including disease and diagnosis, examination, test, surgery, medication, and anatomical site [56]. The entity types for 2021 remained the same as in 2020 [59]. The NER tasks organized by CHIP are more diverse in terms of data types and entity types. In 2020, two NER tasks were released. One of them focused on identifying and extracting entities from Chinese traditional medication instruction texts, consisting of 13 types: drug-related entities, including drug, drug ingredient, disease, symptom, syndrome, disease group, food, food group, person group, drug group, drug dosage, drug taste, and drug efficacy. The other task aimed to identify and extract clinical entities from medical documents, with entities divided into 9 types, including diseases, symptom, drug, medical equipment, procedure, body part, test item, microorganism, and department [65]. In 2023, two NER tasks were released. One task focused on few-shot NER recognition, defining 15 labels, including item, sociology, disease, etiology, body, age, adjuvant, therapy, electroencephalogram, equipment, drug, procedure, treatment, microorganism, department, epidemiology, symptom, and others [70]. Another task focused on identifying PICOS information from Chinese medical literatures, where PICOS elements include population, intervention, comparison, outcome, and study design [73]. Additionally, in 2021, DCIC organized a task that required identifying annotated entities from pathology reports of tumors.

Entity normalization is generally performed after NER. The purpose of entity normalization is to map entities to a unified standard terminology in order to facilitate information exchange and knowledge sharing. In the field of biomedical research, commonly used standard terminologies include International Classification of Diseases 10th Revision (ICD-10), Unified Medical Language System (UMLS), Systematized Nomenclature of Medicine Clinical Terms (SNOMED CT), and others. In 2019, the CHIP organized a entity normalization task with the aim of mapping surgery-related entities in Chinese EHRs to the standard terminology "ICD9-2017 Peking Union Medical College Clinical Edition". In 2020 and 2021, CHIP organized two entity normalization tasks for diagnosis-related entities using the standard terminology "ICD-10 Beijing Clinical Edition v601".

Event extraction refers to the process of identifying specific events or occurrences mentioned in text. Attribute extraction aims to extract specific attributes or features from text. In clinical, it focuses on extracting events or attributes related to healthcare and clinical processes. For example, in the radiology reports of lung cancer and breast cancer, the task defined by CHIP in 2018 included three attributes: tumor size, primary tumor site, and metastatic site. Subsequently, CCKS organized tasks in 2019 and 2020 to further explore the extraction of the same three attributes [55,56]. In 2021, CHIP organized the clinical event extraction task, which aimed to identify four attributes from a given medical history or medical imaging report: anatomical location, subject, description, and occurrence status.

Relation extraction is the process of identifying entities from unstructured texts and determining the relationships between these entities. In 2020, CHIP organized a relation extraction task which defined 53 types of relationships, and required the analysis of medical text sentences to output of all relationships that met the specified conditions [66]. In 2022, CHIP organized another relation extraction task that focused on extracting three key types of medical causal inference relationships from online consultation texts [51]. These relationships include causal relationships, conditional relationships, and hypothetical relationships.

Optical character recognition is a specific type of information extraction task, enabling conversion of printed or handwritten text into machine-readable text. In clinical practice, various paper-based medical documents are generated, and the information contained within them can be used for assist in clinical diagnosis and medical insurance claims [75]. In 2022, CHIP released a task in which the organizer collected scanned images of four types of medical records, including discharge summaries, outpatient invoices, medication invoices, and hospitalization invoices [68]. The task explored the structured data generation and information extraction, which would be utilized for insurance claims. In 2023, CHIP released another task with a corpus of scanned drug package inserts, aiming to identify entities and relationships within them [71].

Some information extraction tasks may contain multiple subtasks. For example, in 2022, CHIP organized a task which aims to extract semantic associations between genes and diseases from scientific literature [53]. This task defined 12 types of named entities, including nine molecular objects and three regulations. It also required recognizing two semantic roles: ThemeOf and CauseOf, as well as four types of regulatory types: Loss of Function, Gain of Function, Regulation, and Compound Change of a Function. In another task released by CHIP in 2022, which aims to extract medical decision trees from unstructured texts such as clinical practice guidelines and medical textbooks [54]. The task required the system to identify entities and relationships within the text, and interconnect the information in order to construct a complete clinical decision-making process.

Information extraction is a foundational and most common task in Chinese biomedical text mining. Transformer-based models such as BERT-wwm, ERNIE and RoBERTa-zh are the mainstream methods. These models can effectively capture the context and semantic meaning of character and words, and have shown significant improvements in performance. In recent tasks, generative language models have begun to show their potential. These generative models are typically combined with extractive models, and their outputs are subsequently fused to final prediction, resulting in generally excellent performance[76,77].

### 3.2. Text classification and text similarity

Text classification refers to the task of categorizing texts in the field of biomedical sciences based on their themes, types, or other characteristics. In 2019, CHIP introduced a clinical trial eligibility criteria text

**Table 2**
Summary of evaluation tasks in China Health Information Processing Conference (CHIP).

| Year | Evaluation task abbreviation | Evaluation task title | Dataset description | Dataset size | Link |
|---|---|---|---|---|---|
| 2023 | PromptCBLUE | CHIP-PromptCBLUE medical large model evaluation | The dataset is sourced from the CBLUE benchmark, encompassing 18 scenarios of medical natural language processing tasks, and includes over 450 instruction fine-tuning templates. | Training set: 87100, validation set: 8456, test set: 8456 | http://cips-chip.org.cn/2023/eval1 |
| 2023 | NER | Chinese medical text few-shot named entity recognition evaluation | The dataset comprises manually annotated entities relevant to medical clinical contexts within medical texts. It includes 15 labels: item, sociology, disease, etiology, body, age, adjuvant, therapy, electroencephalogram, equipment, drug, procedure, treatment, microorganism, department, epidemiology, symptom, and others. | Training set: 400, validation set: 100, test set: 100 | http://cips-chip.org.cn/2023/eval2 |
| 2023 | MedOCR | Drug paper document recognition and entity relation extraction | The drug leaflets were manually annotated for drugs, diseases, and clinical findings. | Training set: 400, validation set: 200, test set: 400 | http://cips-chip.org.cn/2023/eval3 |
| 2023 | YIER-LLM | CHIP-YIER medical large model evaluation task | A series of multiple-choice questions constructed from medical entrance exam questions, clinical practice physician assessments, medical textbooks, medical literature/guidelines, and publicly available medical records. Dataset size: 1500. | Training set: 1000, test set: 500 | http://cips-chip.org.cn/2023/eval4 |
| 2023 | PICOS | Medical literature PICOS identification | The dataset consists of titles and abstracts from medical publications, annotated with five categories: Population (P), Intervention (I), Comparison (C), Outcome (O), and Study Types (S) | Training set: 2500, validation set: 1000, test set: 1000 | http://cips-chip.org.cn/2023/eval5 |
| 2023 | DTC | Chinese diabetes question classification evaluation | The dataset consists of diabetes questions from internet, encompassing six categories: diagnosis, treatment, common knowledge, healthy lifestyle, epidemiology, and others. | Training set: 6000, validation set: 1000, test set: 1000 | http://cips-chip.org.cn/2023/eval6 |
| 2022 | AGAC | Text mining task for gene-disease association semantics | The AGAC corpus comprises 12 categories of molecular entities related to "gene-disease" associations and their triggering term entities: Var, MPA, Interaction, Pathway, CPA, Reg, PosReg, NegReg, Disease, Gene, Protein, and Enzyme. It includes semantic role annotations: ThemeOf and CauseOf; regulatory types: Loss of Function (LOF), Gain of Function (GOF), Regulation (REG), and Composite changes in function (COM). | Training set: 250, test set: 2000 | http://www.cips-chip.org.cn/2022/eval1 |
| 2022 | CMedCausal | Medical causal entity and relation extraction | The annotated dialogue corpus includes three types of relationships: "causal", "conditional", and "hyponymy" | Training set: 2000, test set: 2000 | http://www.cips-chip.org.cn/2022/eval2 |
| 2022 | Text2DT | Extracting medical decision trees from medical texts | The dataset consists of extracted diagnostic and therapeutic decision trees from clinical practice guidelines and medical textbooks. A diagnostic and therapeutic decision tree is defined as a binary tree composed of conditional nodes and decision nodes. | Training set: 300, validation set: 100, test set: 100 | http://www.cips-chip.org.cn/2022/eval3 |
| 2022 | MedOCR | Identification of electronic medical paper documents | The dataset consists of scanned images of paper medical records from the internet, defining 87 attributes to be extracted, which include types such as discharge summaries; outpatient invoices; pharmacy purchase invoices; and hospitalization invoices. | Training set: 1000, validation set: 200, test set: 500 | http://www.cips-chip.org.cn/2022/eval4 |
| 2022 | CDN | Clinical diagnostic coding | The dataset comprises annotated information extracted from EHRs, including diagnostic details (such as admission diagnosis, preoperative diagnosis, postoperative diagnosis, and discharge diagnosis), as well as surgical names, medication names, and medical order names. | Training set: 2700, test set: 337 | https://www.cips-chip.org.cn/2022/eval5 |
| 2021 | MDCFNPC | Classifying positive and negative clinical findings in medical dialog | The data comes from publicly available internet-based telemedicine consultations includes patient chief complaints and physician diagnostic judgments, categorized into four attributes: negative, positive, other, and unspecified. | Training set: 6000, validation set: 2000, test set: 2000 | http://www.cips-chip.org.cn/2021/eval1 |
| 2021 | CDEE | Event extraction of clinical discovery | The dataset extracted present medical history or imaging findings reports from EHRs, involving attributes across four dimensions: anatomical sites, main terms, descriptive terms, and occurrence status. | Training set: 2070, test set: 532 | http://www.cips-chip.org.cn/2021/eval2 |
| 2021 | CDN | Normalization of Chinese clinical terminology | The dataset comprises diagnostic entities, partial surgical entities, and standardized surgical relationship corpora extracted from Chinese EHRs. Dataset size: NA. | Training set: 9699, test set: 801 | http://www.cips-chip.org.cn/2021/eval3 |
| 2020 | CMeEE | Chinese medical text named entity recognition | The dataset comprises 9 entity types extracted through medical text mining: diseases, clinical manifestations, medications, medical devices, medical procedures, anatomical structures, medical laboratory tests, microorganisms, and departments. | Training set: 15000, validation set: 5000, test set: 6618 | http://www.cips-chip.org.cn/2020/eval1 |
| 2020 | CMeIE | Chinese medical text relationship extraction | The pediatric training corpus and the corpus extracted from a hundred common diseases yielded 53 schemas, comprising 10 synonymous relations and 43 other relations. | Training set: 17924, validation set: 4482, test set: 5602 | https://www.cips-chip.org.cn/2020/eval2 |
| 2020 | CDN | Clinical terminology normalization | The dataset includes diagnostic entities extracted from Chinese EHRs. | Training set: 8000, test set: 10,000 | http://www.cips-chip.org.cn/2020/eval3 |
| 2020 | Covid19 | Prediction of epidemic trends in COVID-19 | The dataset comprises regional time-series data of confirmed COVID-19 cases, including daily counts of newly diagnosed cases. | NA | https://www.cips-chip.org.cn/2020/eval4 |
| 2020 | TCM-QA | Question generation of traditional Chinese medicine literature | Texts from the field of Traditional Chinese Medicine (TCM), including four TCM books and selected texts from TCM forums, with manually constructed question–answer pairs. | Training set: 3500, validation set: 750, test set: 750 | https://www.cips-chip.org.cn/2020/eval5 |

Table 2 (*continued*)

| Year | Evaluation task abbreviation | Evaluation task title | Dataset description | Dataset size | Link |
|---|---|---|---|---|---|
| 2020 | TCM-NER | Entity recognition in traditional Chinese medicine instructions | The dataset comprises 13 types of entities extracted from traditional Chinese medicine drug instructions: DRUG, DRUG_INGREDIENT, DISEASE, SYMPTOM, SYNDROME, DISEASE_GROUP, FOOD, FOOD_GROUP, PERSON_GROUP, DRUG_GROUP, DRUG_DOSAGE, DRUG_TASTE, and DRUG_EFFICACY. | Training set: 1200, validation set: 400, test set: 397 | https://www.cips-chip.org.cn/2020/eval6 |
| 2019 | CDN | Clinical terminology normalization | The dataset consists of real surgical entities extracted from Chinese electronic medical records that require standardization. | Training set: 4000, validation set: 1000, test set: 2000 | https://www.cips-chip.org.cn:8000/evaluation |
| 2019 | STS | Disease question-answering based on transfer learning | The dataset comprises extracted online disease question–answer sentence pairs related to diabetes, hypertension, hepatitis, aids, and breast cancer. | Training set: 20000, validation set: 10000, test set: 50,000 | https://www.cips-chip.org.cn:8000/evaluation |
| 2019 | CTC | Text classification of Chinese clinical trials eligibility criteria | Descriptive sentences of Chinese clinical trial inclusion/exclusion criteria, and a predefined set of 44 semantic categories for these criteria. | Training set: 22962, validation set: 7682, test set: 7697 | https://www.cips-chip.org.cn:8000/evaluation |
| 2018 | CNER-AE | Entity and attribute extraction of Chinese electronic medical records | Imaging examination reports related to lung cancer and breast cancer. | Training set: 600, test set: 200 | https://icrc.hitsz.edu.cn/chip2018/Task.html |
| 2018 | STS | Patient health consultation question pairs matching | The dataset comes from real patient health consultation corpus. Given two sentences, it is required to determine whether the intentions are the same or similar. | Training set: 20000, test set: 10,000 | https://icrc.hitsz.edu.cn/chip2018/Task.html |

**Table 3**
Summary of evaluation tasks in Chinese Conference on Information Retrieval (CCIR), Chinese Society of Medical Information (CSMI), China National Conference on Computational Linguistics (CCL), and Digital China Innovation Contest (DCIC).

| Year | Evaluation task abbreviation | Evaluation task title | Dataset description | Dataset size | Link |
|---|---|---|---|---|---|
| 2021 | CNER | Medical entity recognition based on pathology report text | Extracted 10 types of entities from pathological text. | Training set: 1000, test set: 1050 | https://www.datafountain.cn/competitions/498 |
| 2021 | IMCS | Intelligent medical dialogue diagnosis and evaluation | The dataset consists of dialogue cases from online medical consultation platforms, utilized for entity recognition, simulated dialogues, and disease diagnosis. Each sample in the dataset includes disease category, patient self-description text, symptoms, and entities and labels inferred from entire medical dialogues. | >2000 | https://www.fudan-disc.com/sharedtask/imcs21/index.html |
| 2020 | PHQC | Classification of public health questions | The dataset consists of public health queries categorized into six major themes: diagnosis, treatment, anatomy/physiology, epidemiology, healthy lifestyle, and choosing healthcare providers. Dataset size: 8000. | Training set: 5000, test set: 3000 | https://www.heywhale.com/home/competition/5f2d0ea1b4ac2e002c164d82/content |
| 2019 | EMR-Query | Data query-based question answering using electronic health records | The dataset consists of query-based question–answer pairs derived from EHRs. Dataset size: NA. | Training set: 1800, validation set: 600, test set: 600 | https://www.biendata.xyz/competition/ccir2019/ |

classification task, which aimed to classify criteria sentences into 44 predefined semantic categories [28]. In 2021, CHIP released a task which aims to classify clinical descriptions collected from internet into negative and positive categories, based on their relevance to patient conditions. In 2023, CHIP released an internet diabetes consultation question classification task, which defined six classes, including diagnosis, treatment, common knowledge, healthy lifestyle, epidemiology, and other [74]. In 2021, CSMI organized a task focused public health questions, which defined six classes, including diagnosis, treatment, anatomy/physiology, epidemiology, healthy lifestyle, and physician selection. It is worth noting that all of these tasks are single-label classification tasks.

Text similarity refers to the measurement of the semantic or content-related similarity between two texts. For instance, in the tasks released by CHIP in 2018 and 2019, patient health consultation question corpora were collected from the internet, and given two questions, the objective was to determine if their intents were the same. On the other hand, in the task released by CCKS in 2021, focusing on medical popular science knowledge, questions and answers were provided, and the objective was to judge whether they matched or not [61].

For such tasks, commonly employed models include Text Convolutional Neural Networks (Text CNN), Text Recurrent Neural Networks (Text RNN), and BERT. However, with the advancement of large language models, current solutions leveraging models like Baichuan2-13B [78] and Qwen-7B [79] have exhibited markedly superior performance.

### 3.3. Knowledge graph and question answering

The knowledge graph is a structured knowledge representation used to store and organize knowledge. It consists of a network of entities and their relationships. Knowledge graph have extensive applications in the field of biomedicine, it provides a way to organize and integrate diverse biomedical data from various sources, enabling researchers to explore complex relationships and make new discoveries. Biomedical knowledge question answering utilizes the biomedical knowledge graph and NLP techniques to enable computers to answer natural language questions related to biomedicine or healthcare, assisting doctors and researchers in more efficiently accessing and applying biomedical knowledge.

In 2020, CCKS released the task of construction and question-answering of COVID-19 knowledge graph [60]. It defined four sub-tasks, including: 1) inference of entity types in the COVID-19 knowledge graph, 2) prediction of hierarchical relationships between entities, 3) link prediction, such as targeted effects of drugs and viruses or protein interactions, and 4) knowledge question answering: constructing question-answering data focused on specific subjects such as health, medicine, and disease prevention and control related to COVID-19. In 2021, CCKS released the task for link prediction in a multi-level knowledge graph of phenotypes, drugs, and molecules, aims to predict seven relationship categories [61]. In 2019, CCIR organized a task that provided a medical event graph built with EHRs, along with a series of natural language questions, requiring participants to build systems that can automatically return results for the given questions.

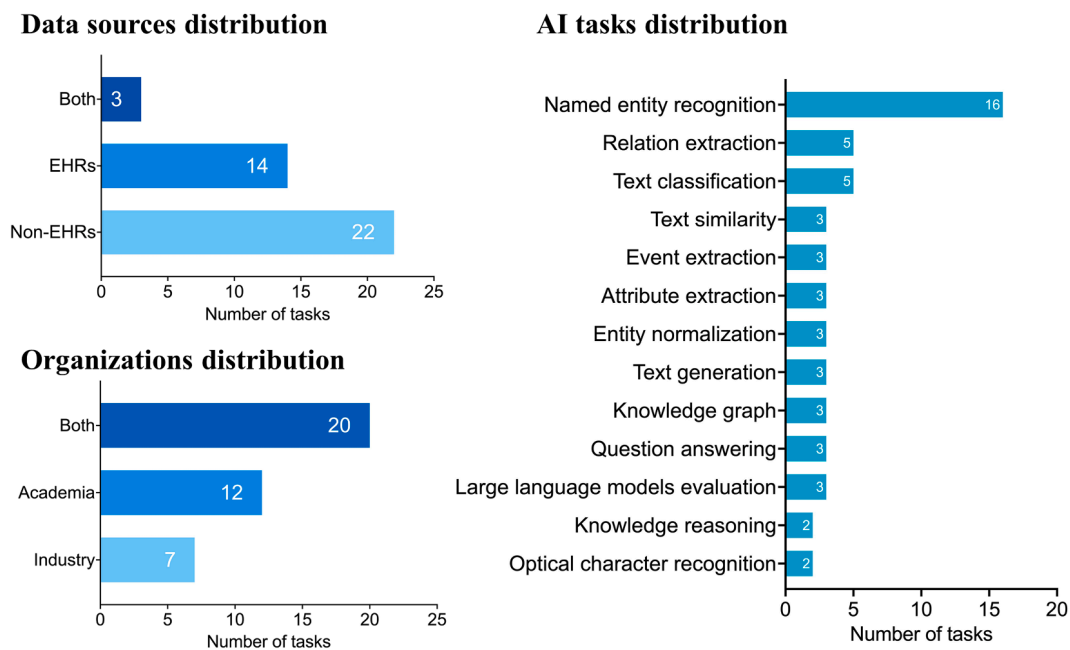

**Fig. 2.** The distribution of data sources, organizations, and artificial intelligence tasks in Chinese biomedical text mining community challenges.

*3.4. Text generation and knowledge reasoning*

Text generation refers to the process of generate natural language text, while knowledge reasoning refers to the process of using existing knowledge and information to make inferences. Both of them are complex and challenging NLP tasks. In 2020, CHIP organized a question generation task focused on Chinese traditional medical literatures and related texts from internet forum. Participants were required to develop algorithms that process these texts and generate questions. In 2021, CCKS organized a dialogue generation task, which was based on the medical dialogue dataset MedDG [80] related to 12 types of common gastrointestinal diseases. The task aimed to generate question–answer pairs containing 160 related entities from five categories: diseases, symptoms, attributes, examinations, and medications. At same year, CCKS organized another task focused on the reading comprehension of Chinese popular medical knowledge. Given texts and questions, the task aims to extract corresponding text spans as answers. For medical knowledge reasoning, in 2021, CCL released the medical dialogue based intelligent diagnosis evaluation task, which explored the identification of medical entities and symptom information from doctor-patient dialogue texts, the automatic generation of medical reports, and the simulation of dialogue process to determine specific diseases. In 2022, CHIP organized a task, which aims to automatic clinical diagnostic coding by given relevant diagnostic information (e.g., admission diagnosis, preoperative diagnosis, postoperative diagnosis, and discharge diagnosis), surgery, medication, and medical advice [52].

*3.5. Large language model evaluation*

Large language models possess powerful capability in text understanding and generation. Extensive research has explored wide range of potential applications for LLM. In biomedical and healthcare fields, higher evaluation standards are required for development and application of LLM due to its specialization, rigor, privacy, and ethical considerations. Ensuring the reliability and credibility of these models in clinical applications is crucial, making comprehensive evaluations of biomedical LLM extremely important. In 2023, based on the benchmark of CBLUE [67], CCKS organized a task which transformed various NLP tasks within different medical scenarios into prompt-based language generation tasks, creating a large-scale prompt tuning benchmark PromptCBLUE[63,64]. The CCKS-PromptCBLUE dataset includes 16 NLP tasks and 94 instruction fine-tuning templates. Next, the organizers optimized the dataset and conducted evaluation tasks at CHIP in 2023 [69]. The CHIP-PromptCBLUE includes 18 NLP tasks and over 450 instruction fine-tuning templates. The PromptCBLUE tasks has two tracks, the parameter-efficient fine-tuning (PEFT) track and in-context learning (ICL) track. In the PEFT track, the Low-Rank Adaptation (LoRA) [81]was the most popular PEFT method to fine-tune the LLM backbone, and the Baichuan-13B-Chat model achieved the highest performance [82]. In the ICL track, the 13B models performed better than 7B models, and the Chinese-LlaMA2-13B-chat achieved the best performance [83].

Additionally, CHIP organized another task in 2023, which released a dataset CHIP-YIER-LLM contains various multiple-choice questions collected from medical licensing exams, medical textbooks, medical literatures, clinical practice guidelines, publicly available EHRs [72]. This task was designed to assess the capabilities of LLM in the field of biomedical research. The state-of-the-art solution used ChatGPT as the evaluation model, and employed prompt learning and ensemble learning techniques [72].

**4. Translational informatics in biomedical text mining**

Translational informatics refers to the application of data science and computational approaches to bridge the gap between biomedical research and clinical practice. It is dedicated to realizing the potential of scientific discovery by accelerating its translation into tangible benefits for patients and healthcare systems. Here we explore the broad potential biomedical applications of these evaluation tasks from the perspective of translational informatics.

As shown in Fig. 3, the community challenges of biomedical text mining first collect, curate and manage large amounts of health data from a wide range of resources. These resources encompass electronic health records, scientific literatures, clinical trial reports, medical books, clinical practice guidelines, and even the internet. The complexity and heterogeneity of these data necessitate the development and deployment of sophisticated NLP technologies to extract valuable information and discover latent knowledge. Through the application of diverse NLP techniques, ranging from named entity recognition, entity normalization, attribute extraction, relation extraction, event extraction, to text classification, text similarity, knowledge graph, question answering, and text generation, biomedical text mining enables the efficient processing and analysis of biomedical textual data. This, in turn, facilitates various clinical translational applications such as medical text mining and information processing, clinical data management, terminology standardization, health consultation, question-answering, disease prediction, epidemic forecasting, clinical decision-making, and knowledge discovery. Ultimately, biomedical text mining not only streamlines the integration of research findings into clinical workflows, but also empowers healthcare professionals with the tools necessary for intelligent medicine and health management.

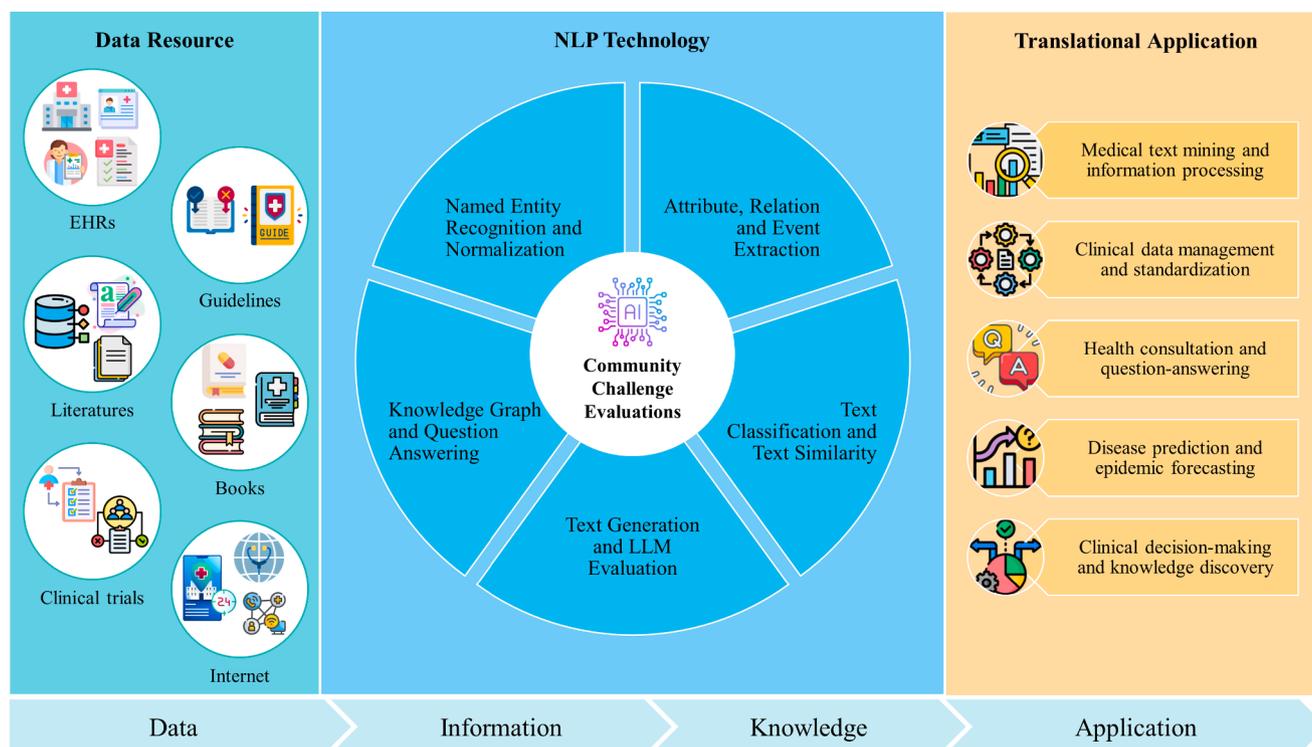

**Fig. 3.** The perspective of translational informatics for biomedical text mining community challenges, which plays a crucial role in health information processing that collected from diverse biomedical resources by leveraging various natural language processing technologies, and has numerous applications in clinical translational research.

## 5. Discussion and perspective

### 5.1. Comparison of Chinese and English BioNLP community challenges

Here we provide a comparison between the Chinese BioNLP evaluation tasks and their commonly used English counterparts. Understanding the similarities and differences is crucial for recognizing the unique challenges and opportunities within the Chinese BioNLP landscape.

First, the BioNLP community challenges in China has started in the past decade, while the international BioNLP community challenges have a history of more than two decades. These global challenges have not only paved the way for global advancements and innovations, but has also provided invaluable insights for the BioNLP community challenges in China.

Second, Chinese is a logographic language without word boundaries, complicating word tokenization and information extraction, whereas English's alphabetic system with clear word boundaries facilitates these processes. Chinese also has a higher degree of ambiguity and polysemy, where characters or words can have multiple meanings depending on context, significantly impacting NLP models' performance in tasks like NER and event extraction. For example, there are some TCM related BioNLP tasks such as question generation in TCM literatures and entity recognition in TCM instructions, which require models to have a deep understanding of TCM, and be able to handle the specialized terms and concepts.

Third, the English BioNLP challenges, such as i2b2, primarily focus on electronic health records [84], while BioCreative focuses on biomedical literature [85]. English datasets also benefit from numerous domain standards, including ontologies like UMLS [86] and SNOMED CT [87], as well as databases like PubMed [88] and MIMIC [89]. In contrast, Chinese BioNLP tasks also encompass a wide range of sources, including electronic medical records, scientific literature, clinical trial reports, medical books, clinical practice guidelines, and even social media data. However, there are not enough domain standards to refer to, which increases the difficulty of data integration and standardization.

Fourth, In English BioNLP, datasets like i2b2, TREC and BioCreative often benefit from well-curated terminology standards and well-established annotation guidelines, which are publicly available and widely used in the research community. Conversely, the annotation of Chinese datasets lacks standardized annotation guidelines and requires the development of more context-sensitive annotation frameworks. The availability of data is also more limited compared to English. However, efforts are being made to improve standardization and consistency [43].

Finally, the English BioNLP community challenges often attract a global audience, and often influence the development of BioNLP tools and systems used worldwide. The Chinese BioNLP community challenges are more localized, and more specifically targeted to the Chinese healthcare context. In addition, the organizers and participants with a medical background in the Chinese BioNLP task are relatively fewer compared to the English BioNLP task. Medical expertise is crucial for understanding practical needs and characteristics in the medical field. A higher proportion of participants have medical expertise will enhance the clinical relevance of the tasks.

*5.2. Contributions of community challenges*

Community challenges has made significant contributions for biomedical text mining research in many aspects. We summarize five notable contributions ranging from research level to application level, including driving the development of data standards and evaluation benchmarks, promoting technological innovation and development, fostering interdisciplinary collaboration and communication, promoting education and training, facilitating translational informatics application (see Fig. 4).

*5.2.1. Datasets and benchmarks*

The community challenges generally release datasets that are related to specific NLP tasks. These datasets are manually annotated by domain experts, and are sourced from a wide range of biomedical materials, such as published literatures, clinical records, and internet forums. These datasets serve as valuable resources for training and evaluating models, as well as benchmarks for comparing the performance of different approaches. For example, researchers have developed evaluation benchmarks such as CBLUE [67] and MedBench [90], which combine datasets from multiple shared tasks and aims to evaluate the capabilities of language models in Chinese biomedical language understanding. Moreover, researchers have also transformed numerous tasks into prompt-based language generation tasks and construct an evaluation benchmark PromptCBLUE [63], which aims to evaluate the capabilities of large language models in healthcare domain.

*5.2.2. Technology innovation and development*

The community challenges encourage participants to propose state-of-the-art algorithms and methodologies to address problems in biomedical text mining. This competitive environment stimulates researchers' innovative thinking and technological advancements, driving the development and breakthroughs in the field. Taking the information extraction tasks as an example. Initially, early solutions relied on rule-based and dictionary-based methods. Then, deep neural network methods, particularly long short-term memory (LSTM) architecture, gradually gained traction and garnered significant attention. Subsequently, BERT renowned for their powerful text representation capabilities, were often used by participants and found applications in various biomedical fields [91]. In recent research, large language models such as ChatGPT have also been leveraged for purposes such as data augmentation or zero-shot information extraction [92], showcasing their potential for advancing the field even further.

*5.2.3. Interdisciplinary collaboration and communication*

Numerous experts with different research background participate in these community challenges. These experts, hailing from both academia and industry, come together through academic conferences to showcase their algorithm architectures, performance outcomes, and explore potential collaboration opportunities. Additionally, the community challenges also provide benefits to participants through the publication of papers on specific task topics. This interdisciplinary collaboration and communication play a crucial role in promoting knowledge sharing and cooperation among different domains, thus driving the development of biomedical text mining.

*5.2.4. Education and training*

The evaluation tasks serve as valuable learning and practical resources for students and beginners in the BioNLP domain. They allow participants to focus on specific tasks, engage in competition, discuss with experts, and improve their skills by evaluating their methods and receiving constructive feedback. After the task is completed, the researchers can further utilize the dataset to evaluate the algorithms and publish articles. Moreover, these shared tasks also provide open access to relevant datasets, which are often used by universities for teaching courses and practical exercises to assess student learning performance. It is worth noting that some datasets explicitly state their restriction to be used only during community challenges and are not made available thereafter.

*5.2.5. Translational informatics applications*

Biomedical text mining has the potential to contribute significantly to the standardization of knowledge. By extracting valuable information from vast volumes of biomedical literature, it becomes possible to establish high-quality knowledge bases and ontologies. These resources, in turn, can offer timely evidence-based information for researchers, clinicians, and policy makers. For example, mining the gene-disease semantic associations from literatures can help researchers in discovering novel gene candidates related to diseases [93], while extracting PICOS information from the literature can contribute to the construction of clinical evidence knowledge bases [94]. Biomedical text mining plays a crucial role in clinical decision support. Through the extraction and normalization of clinical information, large quantities of unstructured clinical textual data can be converted into structured and computable formats. This transformation enables clinicians to access comprehensive and accurate patient information, facilitating clinical diagnosis and treatment recommendations. Moreover, biomedical text mining techniques can also be employed in epidemic trend prediction and risk assessment. By mining extensive clinical textual data, it becomes possible to identify key features and patterns associated with patients, which enabling the implementation of appropriate intervention measures to improve patient health outcomes. Biomedical text mining holds significant potential for commercial applications. In the community challenges, multiples evaluation tasks are organized by company. These tasks typically employ data from real-world healthcare sources and have clear clinical application scenarios, such as patient health consultation, and automated question answering. The successful development of solutions for these tasks can provide opportunities and competitive advantages for businesses.

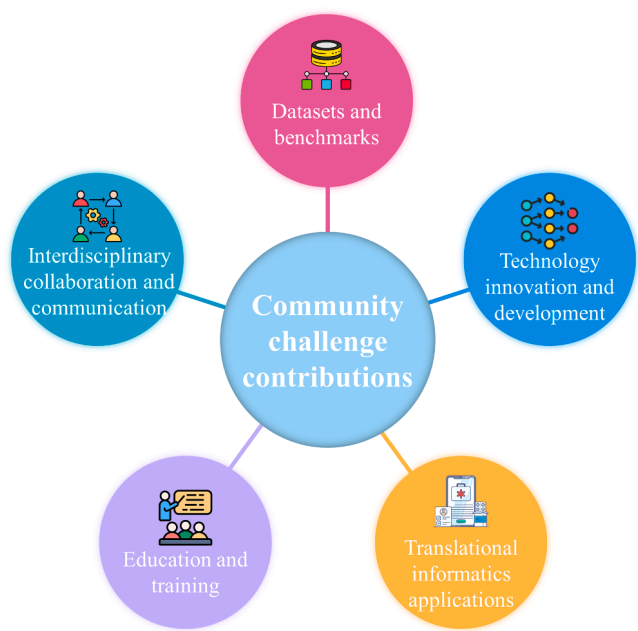

**Fig. 4.** The contributions of community challenges of biomedical text mining.

*5.3. Limitations of current community challenges*

Current community challenge evaluation tasks on Chinese biomedical text mining face several limitations.

Firstly, there is a lack of representativeness of data in evaluation tasks. Biomedical and healthcare data possess sensitivity and privacy concerns, making it difficult to obtain large-scale datasets. This results in many tasks using small or synthetic datasets, which may limit the representativeness and applicability. Data quality and annotation are also challenging, as the annotation of medical data are complex and requires high levels of expertise from annotators. For data types, many Chinese NLP tasks only collect a single type of data. For organizers and participants, there are relatively few people with medical backgrounds. For data availability, some datasets are restricted for use only during the evaluations and not made available afterward, limiting the impact of community challenge evaluation tasks. In the future, task organizer should involve more professionals in medicine background, and should consider incorporating data from multiple sources and modalities.

Secondly, the developed solutions may lack sufficient innovation and reproducibility. Some participants tend to apply established methods to achieve quick results, rather than exploring novel and innovative approaches. This can lead to a lack of innovation and diversity in methods, limiting further technical advancements. On the other hand, some algorithms may perform well on specific datasets but fail to generalize to others, reducing their broad applicability and reliability. To encourage innovation, future community challenges should emphasize the exploration of novel techniques and reward participants for their creativity. When integrating with existing systems, algorithm performance must be evaluated across multiple datasets and scenarios to ensure robustness and generalizability.

Lastly, there exists a translational gap between evaluation tasks and clinical application. Compared with the English BioNLP tasks, such as i2b2 which is very relevant to clinical problems, some Chinese BioNLP tasks focus on basic NLP tasks, such as medical information extraction and text classification, while more intricate scenarios closely aligned with clinical practice are lacking. These tasks are generally abstracted from complex real-world problems and simplified to facilitate evaluation and comparison. However, such simplification may not reflect the complexity, diversity, and ambiguity of problems in real-world clinical settings. In real-world clinical practice, models always need to handle more noise and uncertainty, which may not have been fully considered in current evaluation tasks. Therefore, even systems that perform well in evaluation tasks may not be able to achieve similar results in practical applications.

*5.4. Lessons and guidelines for future community challenges*

The analysis of the Chinese biomedical NLP challenges revealed several key insights that can inform the design and execution of future events. Below are the lessons learned from past challenge outcomes. First, the success of biomedical NLP challenges heavily depends on the quality and diversity of the datasets provided. Challenges that offered well-curated, diverse datasets attracted higher participation and had a more substantial long-term impact. These datasets should represent a wide range of medical domains, document types, and patient populations to ensure that the models developed are robust and generalizable. Furthermore, the open access to challenge datasets and results was found to be vital for promoting reproducibility and broader impact within the research community. Second, the relevance of challenge tasks to real-world clinical needs significantly influenced the outcomes and the eventual application of the developed models. Challenges that addressed specific, unmet needs within healthcare systems—such as tasks involving electronic medical records were particularly valuable. Designing tasks that align closely with current clinical problems enhances the relevance and potential impact of the challenge outcomes. Third, active engagement with the research community through online discussions, workshops, and post-challenge forums was a critical factor in the success of the most impactful challenges. These activities not only fostered collaboration and knowledge sharing but also led to more innovative and effective solutions. The continuous interaction among participants, organizers, and stakeholders contributed to the development of a cohesive and active community that could sustain interest and innovation beyond the challenge period.

Based on these lessons, we have proposed the following guidelines for organizing future biomedical NLP challenges, which will help improve clinical relevance, impact, and sustainability. First, future challenges should emphasize the curation of datasets that are not only large but also diverse in terms of medical domains and data sources. Including real-world clinical data that encompasses a wide range of medical conditions, patient demographics, and document types will enhance the applicability and generalizability of the models developed during the challenge. Open access to these datasets should be maintained to support reproducibility and further research. Second, future challenges should involve clinical stakeholders during the task design phase, to ensure the relevance of challenge tasks to real-world clinical problems. Feedback from clinicians can guide the creation of tasks that address pressing clinical challenges and facilitate the translation of research findings into practice. Additionally, future challenges should explore the evaluation of large language models (LLMs) in real-world clinical scenarios to keep pace with the latest technological advancements. Third, to ensure the continued relevance and success of biomedical NLP challenge series, organizers should plan for their long-term sustainability. Building partnerships with academic institutions, healthcare organizations, and industry can provide the necessary support and resources to sustain the challenge series over time and promote the clinical translation of challenge outcomes.

*5.5. Future perspectives in the era of large language models*

Here, we summarize six perspectives that can enhance the community challenges in biomedical text mining in the era of large language models (see Fig. 5).

*5.5.1. Comprehensive biomedical evaluation benchmark*

With the emergence of large language models such as ChatGPT and its successors, there is an urgent need for comprehensive evaluation of their capability in biomedical text mining or healthcare information processing [95]. Establishing comprehensive benchmarks tailored to Chinese BioNLP is crucial. These benchmarks should reflect the diversity and complexity of Chinese biomedical textual resource, including extensive clinical text data and traditional Chinese medicine resource. These capabilities should include medical information retrieval, medical language understanding, medical text generation, medical knowledge question answering, clinical decision reasoning, and more. To address this, diverse and accessible datasets or benchmarks should be developed, such as PromptCBLUE [63,69], CMB [96], MedBench [90], and MultiMedBench [97].

*5.5.2. Handling multi-source and multimodal data*

Biomedical text data encompasses a wide range of sources, including published literatures, electronic medical records, textbooks, and social media, etc. Additionally, Biomedical data exist in various modalities such as images, videos, and audio. Utilizing these diverse sources and multimodal data can provide more comprehensive information for diagnosis, treatment, and research [97,98]. Therefore, evaluating the model's ability to integrate multiple sources and process multimodal data is crucial in future community challenges. This may involve tasks such as cross-modal information fusion [99], cross-modal querying and retrieval [100].

*5.5.3. Leveraging domain-specific knowledge*

The field of biomedicine has a large number of software, tools, and knowledge bases. Integrating their functionalities into large language models can greatly enhance the models' capabilities. For example, the model can automatically invoke appropriate software and tools to parse, annotate, and analyze biomedical data [101,102]. Moreover, the model can extract relevant information from knowledge bases to generate more accurate responses and reduce erroneous information [103]. Future Chinese BioNLP community challenges should encourage the use of domain-specific knowledge graphs, specialized medical ontologies, and other structured data sources that can enhance the model's understanding of complex medical knowledge.

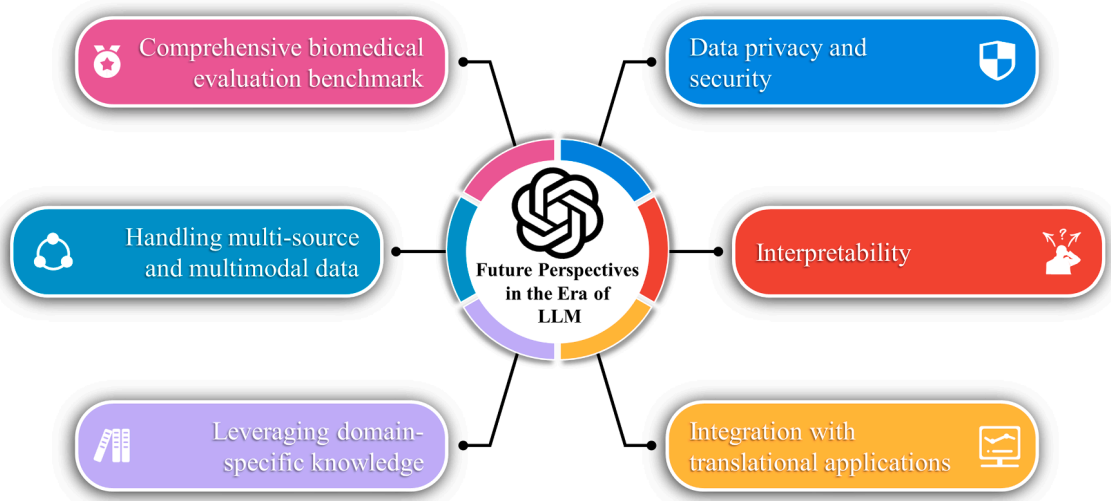

**Fig. 5.** The six future perspectives of community challenges of biomedical text mining in the era of large language models.

*5.5.4. Data privacy and security*

The collection and sharing of biomedical data raise concerns about data privacy and security [104]. It is necessary to consider how to protect sensitive information while performing text mining and information processing using large language models [105,106]. So far, the Chinese biomedical text mining community challenge has not organized any tasks related to data security and privacy protection, no related datasets have been released, and no related evaluation baselines can be referenced. Future community challenges should include evaluation tasks related to data privacy and security to ensure responsible development and deployment of these models.

*5.5.5. Interpretability*

Interpreting and understanding the predictions of models is crucial in clinical practice [107–109]. Reliable explanations and evidence-based support are necessary for clinical decision-making. However, evaluating the interpretability of models represents a significant challenge. This may involve explaining the decision-making process of the model, providing reliable evidence and explanations for the predictions, and effectively interacting with clinical experts to ensure that model's predictions are correctly understood or adjusted based on their knowledge and experience. Tasks like CMedCausal and Text2DT have explored how causal relationships and decision trees in medicine affect clinical decision making. Future challenges should prioritize the organization of evaluation tasks that enhance the transparency and interpretability of model decisions.

*5.5.6. Integration with translational applications*

Applying large language models to clinical practice is an important goal [110]. Evaluating the ability of models to integrate with clinical translational applications is a key challenge. This may involve incorporating models into medical knowledge bases, hospital information systems, clinical decision support systems, or other clinical workflows.

In China, these efforts must be carried out in collaboration with hospitals to achieve further implementation. Therefore, in future Chinese biomedical text mining challenges, participation from hospitals is essential. Only by doing so can we accelerate the translation of research findings into practical healthcare solutions.

# 6. Conclusion

In this review, we investigated the recent advances of community challenges in biomedical text mining in China from a translational informatics perspective. We explored various BioNLP tasks such as named entity recognition, entity normalization, attribute extraction, relation extraction, event extraction, text classification, text similarity, knowledge graph construction, question answering, text generation, and large language model evaluation. We also summarized the potential clinical applications, compared to the English counterparts, discussed the contributions, limitations, lessons and guidelines of these community challenges, and highlighted future directions in the era of large language models.

**Author contributions**

HZ and RW performed data collection, investigation, and analysis. HZ drafted the manuscript. RW revised the manuscript. JC, WF, EW, JL, AS and LT provided constructive suggestion and technique assistance; BS obtained the funding; ZL, BT and BS supervised the study. All authors have read and agreed to the published version of the manuscript.

Statement of Significance

- We identified 39 evaluation tasks from 6 community challenges of Chinese BioNLP spanned from 2017 to 2023.
- We conducted a systematic summary and comparative analysis on these BioNLP tasks, and summarized potential clinical applications from a translational informatics perspective.
- We discussed the contributions, limitations, lessons and guidelines of these community challenges, and highlighted future directions in the era of large language models.
- Community challenges play a crucial role in promoting technology innovation and interdisciplinary collaboration in biomedical text mining and health information processing.
- Community challenges provide platforms for researchers to develop state-of-the-art solutions for various BioNLP tasks.

**CRediT authorship contribution statement**

**Hui Zong:** Writing – original draft, Visualization, Investigation,

Conceptualization. **Rongrong Wu:** Writing – review & editing, Visualization, Investigation, Conceptualization. **Jiaxue Cha:** Data curation, Conceptualization. **Weizhe Feng:** Data curation, Conceptualization. **Erman Wu:** Data curation, Conceptualization. **Jiakun Li:** Data curation, Conceptualization. **Aibin Shao:** Data curation, Conceptualization. **Liang Tao:** Data curation, Conceptualization. **Zuofeng Li:** Supervision, Conceptualization. **Buzhou Tang:** Supervision, Conceptualization. **Bairong Shen:** Supervision, Investigation, Funding acquisition, Conceptualization.

**Declaration of competing interest**

The authors declare that they have no known competing financial interests or personal relationships that could have appeared to influence the work reported in this paper.

**Acknowledgements**

This work was supported by the National Natural Science Foundation of China (Grant numbers 32270690 and 32070671). We also thank these community challenges for providing an open platform, the organizers of evaluation tasks for defining tasks and providing datasets, and the participants of the tasks for developing algorithms or systems.

**References**

[1] L. Shen, et al., The fourth scientific discovery paradigm for precision medicine and healthcare: Challenges ahead, Precis. Clin. Med 4 (2) (2021) 80–84.
[2] Q. Chen, et al., LitCovid in 2022: an information resource for the COVID-19 literature, Nucleic Acids. Res 51 (D1) (2023) D1512–D1518.
[3] Z. Lu, PubMed and beyond: a survey of web tools for searching biomedical literature, Database (Oxford) 2011 (2011) p. baq036.
[4] Y. Yang, et al., Computational modeling for medical data: From data collection to knowledge discovery, The Innovation Life (2024) 100079.
[5] Y. Wang, et al., A knowledge empowered explainable gene ontology fingerprint approach to improve gene functional explication and prediction, iScience 26 (4) (2023) 106356.
[6] Y. Wang, et al., ViMIC: a database of human disease-related virus mutations, integration sites and cis-effects, Nucl. Acids. Res. 50 (D1) (2022) D918–D927.
[7] T. Bekhuis, Conceptual biology, hypothesis discovery, and text mining: Swanson's legacy, Biomed. Digit. Libr 3 (2006) 2.
[8] V. Gopalakrishnan, et al., Towards self-learning based hypotheses generation in biomedical text domain, Bioinformatics 34 (12) (2018) 2103–2115.
[9] J. He, et al., The practical implementation of artificial intelligence technologies in medicine, Nat. Med 25 (1) (2019) 30–36.
[10] E.H. Shortliffe, M.J. Sepulveda, Clinical Decision Support in the Era of Artificial Intelligence, JAMA 320 (21) (2018) 2199–2200.
[11] F. Zhu, et al., Biomedical text mining and its applications in cancer research, J. Biomed. Inform. 46 (2) (2013) 200–211.
[12] P. Przybyla, et al., Text mining resources for the life sciences, Database. (oxford) 2016 (2016).
[13] A. Allot, et al., LitSense: making sense of biomedical literature at sentence level, Nucl. Acids. Res. 47 (W1) (2019) W594–W599.
[14] C.H. Wei, et al., PubTator central: automated concept annotation for biomedical full text articles, Nucl. Acids. Res. 47 (W1) (2019) W587–W593.
[15] S. Zhao, et al., Recent advances in biomedical literature mining, Brief. Bioinform 22 (3) (2021).
[16] C.H. Wei, H.Y. Kao, Z. Lu, PubTator: a web-based text mining tool for assisting biocuration, Nucl. Acids. Res. 41 (Web Server issue) (2013) W518–W522.
[17] R.I. Dogan, R. Leaman, Z. Lu, NCBI disease corpus: a resource for disease name recognition and concept normalization, J. Biomed. Inform. 47 (2014) 1–10.
[18] C.H. Wei, et al., tmVar: a text mining approach for extracting sequence variants in biomedical literature, Bioinformatics 29 (11) (2013) 1433–1439.
[19] C.H. Wei, et al., PubTator 3.0: an AI-powered literature resource for unlocking biomedical knowledge, Nucl. Acids. Res. (2024).
[20] J. Lei, et al., A comprehensive study of named entity recognition in Chinese clinical text, J. Am. Med. Inform. Assoc 21 (5) (2014) 808–814.
[21] X. Yang, et al., Clinical concept extraction using transformers, J. Am. Med. Inform. Assoc 27 (12) (2020) 1935–1942.
[22] Y. Hu, et al., Towards precise PICO extraction from abstracts of randomized controlled trials using a section-specific learning approach, Bioinformatics 39 (9) (2023).
[23] Krallinger, M., et al., The CHEMDNER corpus of chemicals and drugs and its annotation principles. J Cheminform, 2015. 7(Suppl 1 Text mining for chemistry and the CHEMDNER track): p. S2.
[24] L. Luo, et al., BioRED: a rich biomedical relation extraction dataset, Brief. Bioinform 23 (5) (2022).
[25] J. Li, et al., BioCreative V CDR task corpus: a resource for chemical disease relation extraction, Database. (oxford) 2016 (2016).
[26] S. Liu, et al., Drug-Drug Interaction Extraction via Convolutional Neural Networks, Comput. Math. Methods. Med 2016 (2016) 6918381.
[27] J. Chen, et al., Biomedical relation extraction via knowledge-enhanced reading comprehension, BMC. Bioinformatics 23 (1) (2022) 20.
[28] H. Zong, et al., Semantic categorization of Chinese eligibility criteria in clinical trials using machine learning methods, BMC. Med. Inform. Decis. Mak 21 (1) (2021) 128.
[29] Q. Chen, et al., Multi-label classification for biomedical literature: an overview of the BioCreative VII LitCovid Track for COVID-19 literature topic annotations, Database. (Oxford) 2022 (2022).
[30] N. Fiorini, et al., Best Match: New relevance search for PubMed, PLoS. Biol 16 (8) (2018) e2005343.
[31] Y. Chen, et al., Prostate cancer management with lifestyle intervention: From knowledge graph to Chatbot, Clin. Translat. Discovery 2 (1) (2022) e29.
[32] C. Chakraborty, M. Bhattacharya, S.S. Lee, Artificial intelligence enabled ChatGPT and large language models in drug target discovery, drug discovery, and development, Mol. Ther. Nucleic. Acids 33 (2023) 866–868.
[33] M. Malgaroli, et al., Natural language processing for mental health interventions: a systematic review and research framework, Transl. Psychiatry 13 (1) (2023) 309.
[34] S. Liu, et al., SHAPE: A Sample-Adaptive Hierarchical Prediction Network for Medication Recommendation, IEEE. J. Biomed. Health. Inform 27 (12) (2023) 6018–6028.
[35] J. Li, et al., RARPKB: A knowledge-guide decision support platform for personalized robot-assisted surgery in prostate cancer, Int. J. Surg (2024).
[36] M. Liu, et al., Large-scale prediction of adverse drug reactions using chemical, biological, and phenotypic properties of drugs, J. Am. Med. Inform. Assoc 19 (e1) (2012) e28–e35.
[37] Y. Xiong, et al., A unified machine reading comprehension framework for cohort selection, IEEE. J. Biomed. Health. Inform 26 (1) (2022) 379–387.
[38] A. Stubbs, et al., Cohort selection for clinical trials: n2c2 2018 shared task track 1, J. Am. Med. Inform. Assoc 26 (11) (2019) 1163–1171.
[39] Y. Xiong, et al., Cohort selection for clinical trials using hierarchical neural network, J. Am. Med. Inform. Assoc 26 (11) (2019) 1203–1208.
[40] A. Singhal, M. Simmons, Z. Lu, Text mining genotype-phenotype relationships from biomedical literature for database curation and precision medicine, PLoS. Comput. Biol 12 (11) (2016) e1005017.
[41] Y. Tong, et al., ViMRT: a text-mining tool and search engine for automated virus mutation recognition, Bioinformatics 39 (1) (2023).
[42] P.H. Li, et al., pubmedKB: an interactive web server for exploring biomedical entity relations in the biomedical literature, Nucl. Acids. Res. 50 (W1) (2022) W616–W622.
[43] C. Yu, et al., PCAO2: an ontology for integration of prostate cancer associated genotypic, phenotypic and lifestyle data, Brief. Bioinform 25 (3) (2024).
[44] A. Kline, et al., Multimodal machine learning in precision health: A scoping review, NPJ. Digit. Med 5 (1) (2022) 171.
[45] H. Zong, et al., Performance of ChatGPT on Chinese national medical licensing examinations: a five-year examination evaluation study for physicians, pharmacists and nurses, BMC. Med. Educ 24 (1) (2024) 143.
[46] M. Wornow, et al., The shaky foundations of large language models and foundation models for electronic health records, NPJ. Digit. Med 6 (1) (2023) 135.
[47] A.J. Thirunavukarasu, et al., Large language models in medicine, Nat. Med 29 (8) (2023) 1930–1940.
[48] C.C. Huang, Z. Lu, Community challenges in biomedical text mining over 10 years: success, failure and the future, Brief. Bioinform 17 (1) (2016) 132–144.
[49] K. Roberts, et al., Searching for scientific evidence in a pandemic: An overview of TREC-COVID, J. Biomed. Inform 121 (2021) 103865.
[50] D. Mahajan, et al., Overview of the 2022 n2c2 shared task on contextualized medication event extraction in clinical notes, J. Biomed. Inform 144 (2023) 104432.
[51] Li Z, et al. CHIP2022 Shared Task Overview: Medical Causal Entity Relationship Extraction. In: Health Information Processing. Evaluation Track Papers. Singapore: Springer Nature Singapore; 2023.
[52] Luo, G., et al. Overview of CHIP 2022 Shared Task 5: Clinical Diagnostic Coding. in Health Information Processing. Evaluation Track Papers. Singapore: Springer Nature Singapore; 2023.
[53] Ouyang S., et al. Text Mining Task for "Gene-Disease" Association Semantics in CHIP 2022. In: Health Information Processing. Evaluation Track Papers. Singapore: Springer Nature Singapore; 2023.
[54] Zhu W, et al. Extracting decision trees from medical texts: an overview of the Text2DT track in CHIP2022. In: Health Information Processing. Evaluation Track Papers. Singapore: Springer Nature Singapore; 2023.
[55] Han X, et al., Overview of the CCKS 2019 knowledge graph evaluation track: entity, relation, event and QA. arXiv preprint arXiv:2003.03875; 2020.
[56] X. Li, et al., Overview of CCKS 2020 Task 3: named entity recognition and event extraction in Chinese electronic medical records, Data. Intelligence 3 (3) (2021) 376–388.
[57] Y. Xia, Q. Wang, Clinical named entity recognition: ECUST in the CCKS-2017 shared task 2. In CEUR Workshop Proceedings, 2017.
[58] J. Zhang, et al., Overview of CCKS 2018 Task 1: named entity recognition in Chinese electronic medical records. Knowledge Graph and Semantic Computing: Knowledge Computing and Knowledge Understanding: 4th China Conference,

[58] CCKS 2019, Hangzhou, China, August 24–27, 2019, Revised Selected Papers 4, Springer, 2019.
[59] C. Ma, W. Huang, Named Entity recognition and event extraction in chinese electronic medical records, Springer Singapore, Singapore, 2022.
[60] T. Jia, et al., Link prediction based on tensor decomposition for the knowledge graph of COVID-19 antiviral drug, Data. Intelligence 4 (1) (2022) 134–148.
[61] B. Qin, et al., Ccks 2021-evaluation track, Springer, 2022.
[62] Y. Wang, et al., End-to-end pre-trained dialogue system for automatic diagnosis, Springer Singapore, Singapore, 2022.
[63] Zhu W, et al. PromptCBLUE: A Chinese Prompt Tuning Benchmark for the Medical Domain. 2023. arXiv:2310.14151 DOI: 10.48550/arXiv.2310.14151.
[64] Ling H, et al. Advanced PromptCBLUE Performance: A Novel Approach Leveraging Large Language Models. In: Knowledge Graph and Semantic Computing: Knowledge Graph Empowers Artificial General Intelligence. Singapore: Springer Nature Singapore; 2023.
[65] Z. Hongying, et al., Building a pediatric medical corpus: Word segmentation and named entity annotation. Chinese Lexical Semantics: 21st Workshop, CLSW 2020, Hong Kong, China, May 28–30, 2020, Revised Selected Papers 21, Springer, 2021.
[66] T. Guan, et al., CMeIE: Construction and evaluation of Chinese medical information extraction dataset. Natural Language Processing and Chinese Computing: 9th CCF International Conference, NLPCC 2020, Zhengzhou, China, October 14–18, 2020, Proceedings, Part I 9, Springer, 2020.
[67] Zhang N, et al. CBLUE: A Chinese Biomedical Language Understanding Evaluation Benchmark; 2021. arXiv:2106.08087 DOI: 10.48550/arXiv.2106.08087.
[68] L. Liu, et al., Information Extraction of Medical Materials: An Overview of the Track of Medical Materials MedOCR, Springer Nature Singapore, Singapore, 2023.
[69] W. Zhu, et al., Overview of the PromptCBLUE Shared Task in CHIP2023, Springer Nature Singapore, Singapore, 2024.
[70] C. Zhang, et al., CMF-NERD: Chinese Medical Few-Shot Named Entity Recognition Dataset with State-of-the-Art Evaluation, Springer Nature Singapore, Singapore, 2024.
[71] Q. Chen, et al., CHIP 2023 Task Overview: Complex Information and Relation Extraction of Drug-Related Materials, Springer Nature Singapore, Singapore, 2024.
[72] H. Hu, et al., Overview of CHIP2023 Shared Task 4: CHIP-YIER Medical Large Language Model Evaluation, Springer Nature Singapore, Singapore, 2024.
[73] H. Zong, et al., Overview of CHIP 2023 Shared Task 5: Medical Literature PICOS Identification, Springer Nature Singapore, Singapore, 2024.
[74] S. Li, et al., The CHIP 2023 Shared Task 6: Chinese Diabetes Question Classification, Springer Nature Singapore, Singapore, 2024.
[75] M.W. Ma, et al., Extracting laboratory test information from paper-based reports, BMC. Med. Inform. Decis. Mak 23 (1) (2023) 251.
[76] M. Cao, et al., LLM Collaboration PLM Improves Critical Information Extraction Tasks in Medical Articles, Springer Nature Singapore, Singapore, 2024.
[77] Q. Zhang, et al., Task-Specific Model Allocation Medical Papers PICOS Information Extraction, Springer Nature Singapore, Singapore, 2024.
[78] C. Ge, et al., Chinese Diabetes Question Classification Using Large Language Models and Transfer Learning, Springer Nature Singapore, Singapore, 2024.
[79] C. Wu, et al., A Model Ensemble Approach with LLM for Chinese Text Classification, Springer Nature Singapore, Singapore, 2024.
[80] W. Liu, et al., MedDG: An Entity-Centric Medical Consultation Dataset for Entity-Aware Medical Dialogue Generation, in: Natural Language Processing and Chinese Computing: 11th CCF International Conference, NLPCC 2022, Guilin, China, September 24–25, 2022, Proceedings, Part I, Springer-Verlag, Guilin, China, 2022, pp. 447–459.
[81] Hu EJ, et al., Lora: Low-rank adaptation of large language models. arXiv preprint arXiv:2106.09685; 2021.
[82] Ling H., et al. Innovative Design of Large Language Model in the Medical Field Based on chip-PromptCBLUE. In Health Information Processing. Evaluation Track Papers. Singapore: Springer Nature Singapore; 2024.
[83] Liu J, et al. Improving LLM-Based Health Information Extraction with In-Context Learning. In: Health Information Processing. Evaluation Track Papers. Singapore: Springer Nature Singapore; 2024.
[84] Y. Gao, et al., Progress Note Understanding - Assessment and Plan Reasoning: Overview of the 2022 N2C2 Track 3 shared task, J. Biomed. Inform 142 (2023) 104346.
[85] R. Islamaj, et al., The overview of the BioRED (Biomedical Relation Extraction Dataset) track at BioCreative VIII, Database. (oxford) 2024 (2024).
[86] O. Bodenreider, The Unified Medical Language System (UMLS): integrating biomedical terminology, Nucl. Acids. Res. 32 (Database issue) (2004) D267–D270.
[87] D. Lee, et al., Literature review of SNOMED CT use, J. Am. Med. Inform. Assoc 21 (e1) (2014) e11–e19.
[88] Q. Jin, R. Leaman, Z. Lu, PubMed and beyond: biomedical literature search in the age of artificial intelligence, EBioMedicine 100 (2024) 104988.
[89] A.E. Johnson, et al., MIMIC-III, a freely accessible critical care database, Sci. Data 3 (2016) 160035.
[90] Cai Y. et al., Medbench: A large-scale chinese benchmark for evaluating medical large language models. arXiv preprint arXiv:2312.12806; 2023.
[91] Devlin J., et al., Bert: Pre-training of deep bidirectional transformers for language understanding. arXiv preprint arXiv:1810.04805; 2018.
[92] D. Hu, et al., Zero-shot information extraction from radiological reports using ChatGPT, Int. J. Med. Inform 183 (2024) 105321.
[93] J. Pinero, et al., The DisGeNET knowledge platform for disease genomics: 2019 update, Nucl. Acids. Res. 48 (D1) (2020) D845–D855.
[94] I.J. Marshall, et al., Trialstreamer: A living, automatically updated database of clinical trial reports, J. Am. Med. Inform. Assoc 27 (12) (2020) 1903–1912.
[95] A. Arora, et al., The value of standards for health datasets in artificial intelligence-based applications, Nat. Med 29 (11) (2023) 2929–2938.
[96] Wang X, et al., Cmb: A comprehensive medical benchmark in chinese. arXiv preprint arXiv:2308.08833, 2023.
[97] T. Tu, et al., Towards Generalist Biomedical AI, NEJM AI 1 (3) (2024) p. AIoa2300138.
[98] H. Poon, Multimodal Generative AI for Precision Health. NEJM AI Sponsored. 0 (0).
[99] S.R. Stahlschmidt, B. Ulfenborg, J. Synnergren, Multimodal deep learning for biomedical data fusion: a review, Brief. Bioinform 23 (2) (2022).
[100] P. Kaur, H.S. Pannu, A.K. Malhi, Comparative analysis on cross-modal information retrieval: A review, Comput. Sci. Rev 39 (2021) 100336.
[101] T. Schick, et al., Toolformer: Language models can teach themselves to use tools, Adv. Neural. Informat. Process. Syst 36 (2024).
[102] Q. Jin, et al., GeneGPT: augmenting large language models with domain tools for improved access to biomedical information, Bioinformatics 40 (2) (2024).
[103] C. Zakka, et al., Almanac — Retrieval-Augmented Language Models for Clinical Medicine, NEJM. AI 1 (2) (2024) p. AIoa2300068.
[104] W.N. Price 2nd, I.G. Cohen, Privacy in the age of medical big data, Nat. Med 25 (1) (2019) 37–43.
[105] Y. Chen, P. Esmaeilzadeh, Generative AI in medical practice: in-depth exploration of privacy and security challenges, J. Med. Internet. Res 26 (2024) e53008.
[106] J. Pool, M. Indulska, S. Sadiq, Large language models and generative AI in telehealth: a responsible use lens, J. Am. Med. Inform. Assoc (2024).
[107] T. Savage, et al., Diagnostic reasoning prompts reveal the potential for large language model interpretability in medicine, NPJ. Digit. Med 7 (1) (2024) 20.
[108] D.W. Joyce, et al., Explainable artificial intelligence for mental health through transparency and interpretability for understandability, NPJ. Digit. Med 6 (1) (2023) 6.
[109] M.R. Karim, et al., Explainable AI for bioinformatics: methods, tools and applications, Brief. Bioinform 24 (5) (2023).
[110] I.S. Kohane, Injecting artificial intelligence into medicine, NEJM. AI 1 (1) (2024) p. AIe2300197.